# A Novel Symbolic Type Neural Network Model— Application to River Flow Forecasting

George S. Eskander, and Amir F. Atiya, Senior Member, IEEE.

*Abstract*— In this paper we introduce a new symbolic type neural tree network called symbolic function network (SFN) that is based on using elementary functions to model systems in a symbolic form. The proposed formulation permits feature selection, functional selection, and flexible structure. We applied this model on the River Flow forecasting problem.  The results found to be superior in both fitness and sparsity.

## I. INTRODUCTION

In this paper we propose a novel formulation that is more flexible than the traditional models. The proposed model, called symbolic function network (SFN in short), is based on combination of some elementary functions together and adapting their parameters to achieve as best fitting performance as possible for the training data.

In the proposed approach we model the function in the form of a tree. The tree formulation is suitable for our approach because it allows breadth (for example by a sum of many elementary functions) as well as depth in modeling (by concatenations of many functions). Moreover, the tree formulation is relevant to the area of symbolic algebra where the formulas are represented in a tree structure and many manipulation and simplification operations can be done on them. So, using this formulation in system Modeling makes it suitable to apply some of the symbolic algebra approaches to simplify and manipulate the constructed models.  In the proposed framework we develop an algorithm to construct the tree and a tree propagation approach that determines the parameters of the functional forms in the tree. The advantage of the proposed approach is that it lets the system choose the most suitable functional forms and the relations between them. As a result the proposed algorithm is expected to potentially produce more concise functional fits.

There are some works found in literature that model systems in the form of tree.  However, the models most related to this work are the following two models that were recently developed and simultaneous to the work presented here. Chen et al [1] presented the Evolving Additive Tree model (EAT).  This model is also a tree structured hybrid model of Mathematical operations, linear and nonlinear terms. The structure and weights of the additive tree are evolved by a tree structure based evolutionary algorithm, and a random search algorithm, respectively. The other model is the Flexible Neural Tree model (FNT), developed by Chen et al [2]-[7]. It is a kind of irregular multi-layer network that has a tree structure and is being evolved based on a pre-defined instruction/operator set. This allows input selection, and different activation functions. The work proposed in this paper has distinct differences from EAT and FNT models. While the architectures have similarities, the proposed construction algorithms and the parameter determination algorithm are different. The construction algorithms proposed here are based on the concepts of forward greedy and backward greedy search approaches. These are concepts that have been of wide use in the feature selection area [8]-[10]. Concerning parameter determination, we propose a steepest-descent based algorithm, and derive "tree-propagation" approach to determine the gradient.

The paper is organized as follows. The next section describes the proposed method. In Section III we apply the proposed model on the River Flow forecasting application, followed by the conclusion section.

## II. THE PROPOSED METHOD

### a) Overview

Representing a function in a symbolic form in terms of a number of elementary functions (for example powers, exponential functions, sinusoids, etc) and elementary operations (for example +, -, *, /) in standard computer algebra work is typically accomplished in the form of a tree representation [11]. Taking cue from these approaches, our proposed model is in the form of a tree that is built in a constructive way in a top down fashion. Elementary functions are added to the tree one by one in some way so as to achieve as best fitting performance as possible for the given training data. Figure 1 shows an example of a constructed tree. Every node represents some elementary function applied to the sum of variables/functions associated with its child nodes. Each terminal node represents some input variable. By having several layers of the tree several levels of function concatenations can be achieved.

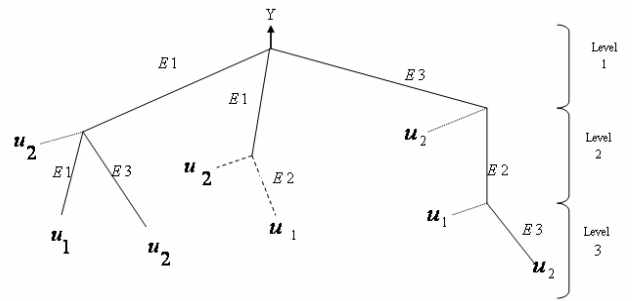

Figure 1 Example of a symbolic function tree

We have considered three basic elementary functions: powers, the exponential function and the logarithm. Since $\log(u)$ is not defined for non-positive $u$ and since $u^v$ might not be defined for negative $u$ or for $u = 0$ for some values of $v$, we have used modifications of these functions to avoid getting trapped in non-defined regions. The final three elementary functions that we use are the following:

$$E1(u) = w(u^2+1)^v$$
$$E2(u) = qe^{\alpha u} \quad (1)$$
$$E3(u) = p\log(u^2+1)$$

Where $u$ is the input argument and $w$, $v$, $q$, $\alpha$, and $p$ are parameters or weights.

Consider again Figure 1, and consider how to interpret the tree. Let $u_1, u_2, \ldots$ be the input variables. Each leaf node denotes an input variable and the link above it gives the elementary function to be applied to it (such as E1, E2, or E3 of Eq. 1). The function modeled is given by (ignore for now the dashed links in the figure):

$$Y = E1(u_2 + E1(u_1) + E3(u_2)) + E1(u_2) + E3(u_2 + E2(u_1 + E3(u_2))).$$

Training is performed in an incremental fashion. For example assume that during the training process we consider adding the dashed links in the figure. After adding the links the network function will be

$$Y = E1(u_2 + E1(u_1) + E3(u_2)) + E1(u_2 + E2(u_1)) + E3(u_2 + E2(u_1 + E3(u_2))).$$

Let the original error before adding the link be $E_{old}$. After adding the link we apply the steepest descent algorithm and adjust all the network weights (all parameters $w$, $v$, $q$, $\alpha$, and $p$ for all nodes), not just the ones that correspond to the added link. Let the new error be $E_{new}$. If $E_{new} < E_{old} - a$ where $a$ is a positive threshold then we keep this added link. Otherwise, we discard it and go back to the old network configuration.

Note that for every link there is a baseline input variable on which it operates. For example at the parent node where we added the new link it used to be $E1(u_2)$, i.e. the baseline variable is $u_2$. After adding the link the baseline variable is kept in addition to the added link, so that $E1(u_2)$ is replaced by $E1(u_2 + E2(u_1))$. The reason for that is that by adding a link we do not want to "disrupt" much of the overall function of the network. By keeping $u_2$ it is like the old function plus some added term. At least theoretically if the multiplier weight $w$, $p$, or $q$ of the new link is zero then we get the same error performance as the old network, thus having possibly a smooth transition when adding the link.

### b) The tree propagation approach

The proposed training algorithm is based on the steepest descent concept. We therefore need to compute the gradient of the error w.r.t. the network weights

In the following we present the proposed algorithm for computation of the gradient. For illustration consider Figure 2.

Consider a particular path along the tree:
Y- Z1- Z2- Z3- ---- where the functions encountered along the path are: $E_{i_1} - E_{i_2} - E_{i_3} -----$

Then, $z_{j-1} = E_{i_j}(z_j) + E_{oth}(z_{oth})$ (2)

where $E_{oth}(z_{oth})$ is the sum of other transformations that are affecting $z_{oth}$ which are the brothers of $z_j$.

Let the error function be

$$J = \sum_{m=1}^{M}(y(m)-d(m))^2 = \sum_{m=1}^{M}e(m)^2 \quad (3)$$

Where $y(m)$ and $d(m)$ are the actual network output and the target output for training example $(m)$ respectively, and $M$ is the size of the training set.

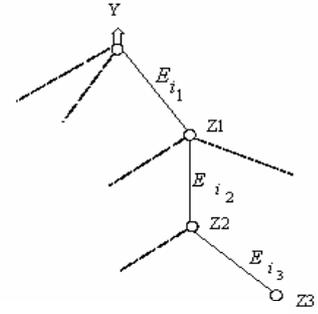

**Figure 2 Particular Path along the tree**

The instantaneous gradient w.r.t some weight $\omega$ is

$$\frac{\delta J}{\delta \omega}(m) = 2\,e(m)\,\frac{\delta y(m)}{\delta \omega} \quad (4)$$

However the total gradient for the whole training set is

$$\frac{\delta J}{\delta \omega} = 2\sum_{m=1}^{M}e(m)\,\frac{\delta y(m)}{\delta \omega} \quad (5)$$

To obtain $\frac{\delta y}{\delta \omega}$ (we skipped the index $m$ for ease notations), a key quantity has to be evaluated, namely $\frac{\delta y}{\delta z_j}$. It is obtained using the chain rule by starting up the tree with ($j=1$), and tracing the tree going downward as follows:

$$\frac{\delta y}{\delta z_j} = \frac{\delta y}{\delta z_{j-1}}\,\frac{\delta z_{j-1}}{\delta z_j} \quad (6)$$

This partial derivative of the network output w.r.t any hidden node output can then be evaluated in a recursive manner. Using (Eq 2) and (Eq 6) we get:

$$\frac{\delta y}{\delta z_j} = \frac{\delta y}{\delta z_{j-1}}\,E'_{i_j}(z_j) \quad (7)$$

Where $E'_{i_j}(z_j)$ is obtained by differentiating the basic functional forms (Eq 1) w.r.t its argument $u$ and evaluated at $u=z_j$.

Once $\frac{\delta y}{\delta z_j}$'s are evaluated for all the tree, the gradient can be obtained.

Let $\omega$ be the weight associated with the node, i.e.

$$z_j = E_{i_{j+1}}(z_{j+1},\omega) + E_{oth}(z_{oth}) \quad (8)$$

$$\frac{\delta y}{\delta \omega} = \frac{\delta y}{\delta z_j} \frac{\delta E_{i_{j+1}}}{\delta \omega} \quad (9)$$

$\frac{\delta E_{i_{j+1}}}{\delta \omega}$ Can be evaluated by differentiating the basic functional forms (Eq.1) w.r.t the weights $w, v, q, \alpha, p$ whatever $\omega$ represents.

### c) Structure Optimization

Structure optimization deals with the issue of constructing the tree; that is determining the strategy and sequence of node additions. We have proposed several algorithms for structure optimization. These algorithms are described as follows:

#### c.1) Forward Algorithm

In the forward algorithm, the network components are added to the network in an incremental way. The network component can be a single link or a complete layer. The algorithm builds the network incrementally from top to down. It starts with an empty network then adds the network elements one by one. After adding each element and training the network it measures the network performance and decides based on that to keep this added item or to restore the previous network configurations..

**Forward Model with Reduced Random Set Capability**

When the number of features is too large, then each layer in the network would grow by too much as we go down the tree. To limit the resulting computational burden, we have proposed here the following variant. Instead of going through all the combinations one by one, at each step we randomly select distinct links from the available complete set. We have introduced a parameter called a reduction factor 'RF' that controls the number of links to be selected at each layer.

#### c.2) Backward Algorithm

For some applications that involve high nonlinearties it is often not practical to train the network in a forward greedy approach by adding a single link each time. The other option is to apply the forward algorithm by adding a complete layer each time. This scenario has an advantage that the network is being constructed and approximates the target in a short time, but it has a disadvantage that the constructed network is not a sparse one and probably contains many superfluous links. For this reason we have designed the backward model. It jumps to a good approximating but not sparse network in a short time- using the forward model by adding a layer each time- but then it applies a pruning algorithm in order to remove the redundant weights from the network.

#### c.3) Forward Backward Algorithm

By applying the forward algorithm on some regression applications we noticed that often while the training error is getting better by adding more links, the generalization performance is getting worst. This is due to the well-known "overfitting" effect. So, we have designed a model variant that is based on running a pruning algorithm in parallel with the constructive algorithm. We found that this scenario enhances the generalization performance and solves the problem of "training error- generalization error" trade off. Also, by applying pruning in parallel there is a chance to re-evaluate the already admitted links.

## III. RIVER FLOW FORECASTING

### a) Introduction

Forecasting of river flow is very important because it can help in predicting agricultural water supply, and flood damage. These predictions can help in agricultural water management and in protection from water shortages and possible flood damage. In this work we used the proposed model to design some predictors of the River Nile flow. Forecasting The Flow of the river Nile can help in determining the optimum amount of water to be released by The High Dam of Aswan (located South of Egypt) so as to optimally fill agricultural and Electricity generating needs

We found that much work have been done on the River Nile flow time series prediction (For a comprehensive survey see [14]). The problem has been treated using linear techniques, such as AR, and ARMA models, and also using nonlinear regression. Among all tried methods, the Multilayer Perceptron Neural Network (MLP) is considered the most reliable forecasting tool as it found to be outperforming all of the linear techniques. Because of the flexibility of the proposed model, we expect that it will give better performance than that of the other traditional neural networks models.

### b) Simulations setup

We used the readings of the average daily River flow volume in millions of cubic meters during the ten years period from 1985 to 1994. This flow is measured at the Dongola station, located in Northern Sudan (South of high Dam). We found that most of the forecasting work done considered one-day ahead forecast; some work had considered longer term prediction such as ten-day or a month ahead. In this work we consider the two prediction problems: ten-day ahead and one-month ahead prediction. Also, for each prediction problem, it could be applied to predict the next step which is called: "single-step ahead prediction", or to predict a multiple step further which is called: "Multi-step ahead prediction". The latter type of prediction is harder. In this work, we tried both single-step and multi-step ahead prediction.

We created two new time series from the original one, the first is ten-day average time series by taking the average of each ten days flow, and the second is one-month average time series by taking the average of the month flow. All time series are scaled before used in the forecasting process and then the output of the forecasting model is being rescaled to get the predicted flow volume. The original time series consists of 3600 points, this is the daily average flow time series in the period of 1985: 1994; taking the ten days average results in a 360 points time series. We divided this time series into two data sets, the first is a training set and consists of 180 points represents the period of 1985:1989, and is used for training the SFN network.

The second set is the testing set which consists of the remaining 180 points representing the period 1990:1994, and is used to test the constructed network.

All versions of the proposed method: 1) Forward Layer by Layer (FLY-SFN), 2) Forward Link by Link (FLK-SFN), 3) Backward (B-SFN), 4) Forward Backward (FB-SFN), and 5) Forward with Reduced Random Set (FRS-SFN) have been tested. The data points are partitioned into 75% training set and 25% validation set. However, the test set is for the final test of the performance of the models. The training set is used for the weight optimization process and the validation set is used to evaluate the network structure in the link admission process. To obtain a comparative idea about the performance of the proposed model, we have implemented on these same problems a multilayer Perceptron neural network (MLP) (a single hidden layer network). We have considered the following methods for training the MLP:

1) The basic Backpropagation (B-BP): It uses the standard gradient descent with the momentum term.

2) The early stopping Backpropagation (ES-BP). It is similar to B-BP in training details, except that the validation data – besides being used in the structure selection- is used in deciding when to stop training.

3) Bayesian regularization Backpropagation (BR-BP) [12]. In this approach the cost function is the error function plus a regularization term that penalizes network complexity. This penalty term is based on a Bayesian formulation.

For all the three training algorithms we used the epoch update mode and set the maximum number of epochs to 10,000. Exploratory runs have been performed to find the best learning rate ($\eta$) and the momentum constant ($\alpha$), and we found that the best values over several tested problems are generally 0.05, and 0.2 respectively. So we fixed the learning rate and momentum constant at these Values for all tested problems. To average out the fluctuations due to the random choice of the initial weights, we have performed 5 runs for every method, each run using different initial weights. The best, worst, and average training and testing performances are reported. For the SFN networks (except for the network constructed using FRS-SFN algorithm) all runs typically lead to the same network (and same performance). In such a case only one run's result is reported. The error measure used to assess the networks performance is Mean Square Error (MSE); where

$$MSE = \frac{\sum_{m=1}^{M}(y(m)-d(m))^2}{M} \; ; y(m), d(m)$$

are the network output, and the desired output at any sample "$m$" respectively; and, $M$ is the length of the investigated data set. Besides the error measures, the number of resulted networks' weights are reported to compare there complexity. In the simulation results, we used the abbreviations of the learning methods followed by the number of hidden nodes for the trained MLP network, and by the maximum network depth for the SFN networks.

### c) Simulations results

#### c.1) Single step ahead prediction

In this experiment, we tried to train the SFN using various algorithms and compared the results with the performance of a single layer MLP. Table I shows the comparison results. Fig 3 shows the SFN network testing performance for the B-SFN (1) (i.e. a single layer backward-algorithm SFN). The results show that: FRS-SFN method results in the sparsest network with the best performance that outperforms all of the other methods.
As shown in the results, that the prediction is accurate and all SFN variants outperform the MLP performance, and also with sparser constructed networks.

#### c.2) Multi step ahead prediction

In addition of predicting just the next step of the time series, we considered here the problem of predicting several steps away, in other words, we are to design a network that gets the available values of the time series as inputs: $x(1), x(2),...x(t),$ and predicts the time series value $x(t+k);$ where $k>1$. surely, the larger k is, the more difficult problem is. We tried to design SFN networks for $k=2,$ and $k=3$. Table II, and Table III, show the comparison results for the double-step and the triple-step predictions respectively. Fig 4, and Fig 5 show the SFN network testing performance for the B-SFN (1) (i.e. a single layer backward-algorithm SFN) for the double-step and the triple-step predictions respectively. As shown in the results, SFN gives the best Prediction accuracy than the resulted MLP networks. For example, while the results of B-BP algorithm in MSE are 0.158, 0.271, and 0.447 for a single, double, and triple Steps ahead respectively; the results of the B-SFN algorithm in MSE are 0.0975, 0.247, and 0.436. Moreover, the SFN constructed networks are sparser than the resulted MLP networks. For example, while the number of weights of the SFN networks constructed using B-SFN algorithm are 9, 6, and 7 for a single, double, and triple Steps ahead respectively; the results of the MLP network constructed using the B-BP algorithm are 46, 31, and 16

TABLE I SINGLE STEP AHEAD PREDICTION (TEN-DAYS AHEAD) OF RIVER FLOW

| Algorithm | Training | | | Testing | | | Average # Weights |
|---|---|---|---|---|---|---|---|
| | Best MSE | Worst MSE | Average MSE | Best MSE | Worst MSE | Average MSE | |
| B-BP (9) | 0.0456 | 0.0837 | 0.0685 | 0.113 | 0.272 | 0.158 | 46 |
| ES-BP (15) | 0.0806 | 0.113 | 0.088 | 0.116 | 0.115 | 0.128 | 76 |
| BR-BP (3) | 0.102 | 0.105 | 0.103 | 0.0858 | 0.147 | 0.113 | 16 |
| FLY-SFN (1) | | 0.115 | | | 0.107 | | 15 |
| FLK-SFN (1) | | 0.129 | | | 0.0937 | | 9 |
| B-SFN (1) | | 0.122 | | | 0.0975 | | 9 |
| FB-SFN (1)(K=5) | | 0.133 | | | 0.0987 | | 7 |
| FRS-SFN(1)(RF=0.5) | 0.127 | 0.147 | 0.137 | 0.0837 | 0.0940 | 0.0884 | 5 |

TABLE II TWO STEP AHEAD PREDICTION (DOUPLE TEN-DAYS AHEAD) OF RIVER FLOW

| Algorithm | Training | | | Testing | | | Average # Weights |
|---|---|---|---|---|---|---|---|
| | Best MSE | Worst MSE | Average MSE | Best MSE | Worst MSE | Average MSE | |
| B-BP (6) | 0.191 | 0.216 | 0.201 | 0.210 | 0.308 | 0.271 | 31 |
| ES- BP (15) | 0.203 | 0.302 | 0.231 | 0.207 | 0.262 | 0.231 | 76 |
| BR –BP (9) | 0.051 | 0.606 | 0.167 | 0.496 | 1.419 | 0.803 | 46 |
| FLY-SFN (1) | | 0.357 | | | 0.253 | | 15 |
| FLK-SFN (1) | | 0.373 | | | 0.248 | | 12 |
| B-SFN (1) | | 0.359 | | | 0.247 | | 6 |
| FB-SFN (1)(K=5) | | 0.272 | | | 0.213 | | 9 |
| FRS-SFN(1)(RF=0.5) | 0.370 | 0.400 | 0.392 | 0.245 | 0.258 | 0.253 | 6 |

TABLE III
THREE STEP AHEAD PREDICTION (TRIPLE TEN-DAYS AHEAD) OF RIVER FLOW

| Algorithm | Training | | | Testing | | | Average # Weights |
|---|---|---|---|---|---|---|---|
| | Best MSE | Worst MSE | Average MSE | Best MSE | Worst MSE | Average MSE | |
| B-BP (3) | 0.312 | 0.542 | 0.429 | 0.386 | 0.513 | 0.447 | 16 |
| ES- BP (12) | 0.298 | 0.375 | 0.335 | 0.344 | 0.388 | 0.361 | 61 |
| BR –BP (3) | 0.296 | 0.316 | 0.306 | 0.425 | 0.450 | 0.440 | 16 |
| FLY-SFN (1) | | 0.582 | | | 0.439 | | 15 |
| FLK-SFN (1) | | 0.542 | | | 0.446 | | 8 |
| B-SFN (1) | | 0.581 | | | 0.436 | | 7 |
| FB-SFN (1)(K=5) | | 0.484 | | | 0.389 | | 12 |
| FRS-SFN(1)(RF=0.5) | 0.617 | 0.618 | 0.618 | 0.439 | 0.444 | 0.441 | 6 |

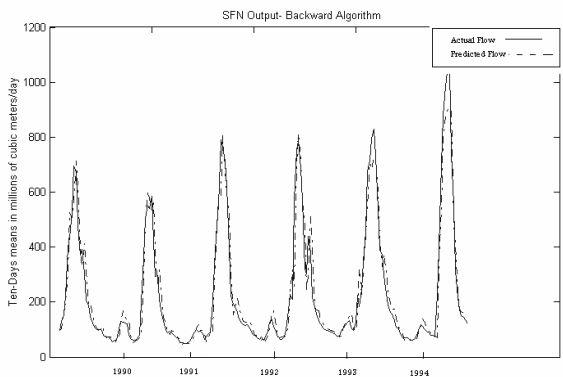

**Figure 3 SFN Network output for the single step ahead prediction (Ten-Days ahead) of river flow**

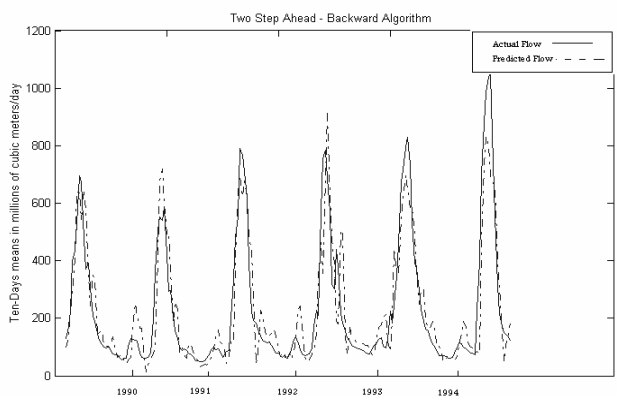

**Figure 4 Two step ahead prediction "double ten-days ahead"**

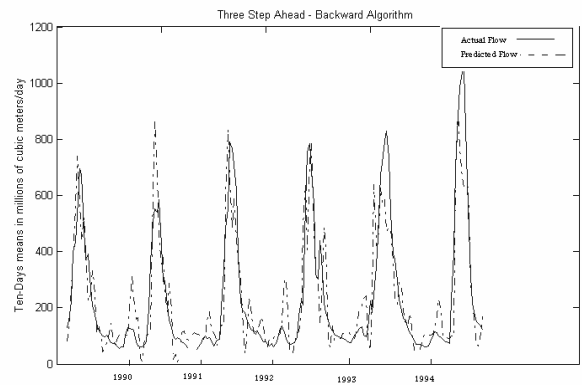

**Figure 5 Three step ahead prediction "double ten-days**

#### c.3) Longer Horizon Forecasting

Instead of predicting several 'k' steps ahead, it could be done by predicting a single step ahead but for a longer horizon step. By other means, we can design a network that predicts the average value of the next k steps instead of predicting 'k' steps ahead value. Theoretically, it is expected that longer horizon forecast could give a comparable performance; in general, averaging cancels the errors. To make sure of this, we designed a network that predicts the average flow of one month ahead; the network performance could be compared by the average of the above three networks performances. As shown in the results of Table IV and Figure 6, predicting the average flow of the next month as a single step ahead prediction of the monthly average flow time series, would give the same performance as predicting the average flow of the next month by taking the average of the three ten-day's intervals predictions. That could be noticed by taking the average performance in MSE of the networks as in tables Table I, II, and Table III and comparing with the performance of the networks as in table IV. For example, the average performance of the SFN-B algorithm for the three single ten days ahead prediction problem is about 0.2601, while the performance of the SFN-B algorithm for a single month ahead prediction problem is 0.261

## IV CONCLUSION

In this paper, a novel symbolic tree based model is introduced. The goal of this model is to synthesize a function made from elementary functions/operations that models a given set of data points in a regression framework. A tree propagation approach is derived to compute the gradients in a backward fashion, and used to design a steepest-descent based optimization algorithm. Algorithms are designed to construct the tree based on the concepts of forward greedy search and backward greedy search. Also, we tested the SFN model as a River Flow time series forecasting tool; and found that it gives better results than the related work found in the literature. In addition, the resulted networks are sparser than the traditional neural networks. Also, we found that predicting a single step ahead is reasonable even for large horizons prediction problems.

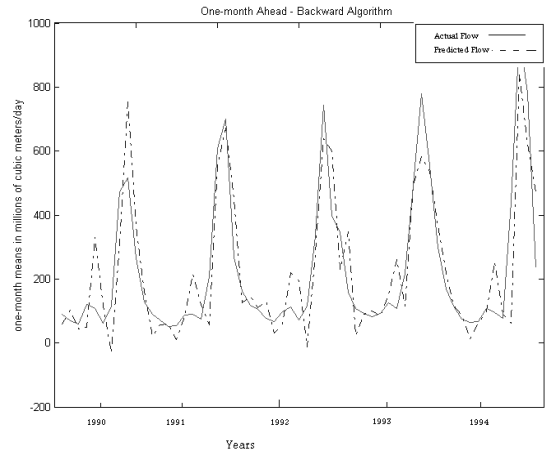

**Figure 6 Single step ahead prediction "one Month ahead"**

TABLE.IV
**SINGLE STEP AHEAD PREDICTION (ONE MONTH AHEAD) OF RIVER FLOW**

| Algorithm | Training | | | Testing | | | Average # Weights |
|---|---|---|---|---|---|---|---|
| | Best MSE | Worst MSE | Average MSE | Best MSE | Worst MSE | Average MSE | |
| B-BP (6) | 0.0398 | 0.0823 | 0.0537 | 0.3769 | 0.5264 | 0.4294 | 31 |
| ES- BP (6) | 0.2265 | 1.3151 | 0.9084 | 0.2637 | 0.9658 | 0.7113 | 31 |
| BR –BP (3) | 0.1117 | 0.1117 | 0.1117 | 0.3411 | 0.3411 | 0.3411 | 16 |
| FLY-SFN (1) | | 0.251 | | | 0.2676 | | 15 |
| LK-SFN (1) | | 0.435 | | | 0.356 | | 7 |
| B-SFN (1) | | 0.319 | | | 0.261 | | 12 |
| FB-SFN (1)(K=5) | | 0.435 | | | 0.356 | | 7 |
| FRS-SFN(1)(RF=0.5) | 0.4312 | 0.4621 | 0.4407 | 0.3556 | 0.3809 | 0.3626 | 4 |